\documentclass[letterpaper,twocolumn,fleqn]{article}
\usepackage{ist}

\makeatletter
\renewcommand{\section}{\@startsection{section}{1}{0mm}%
  {1.2\baselineskip}{0.6\baselineskip}%
  {\normalfont\fontsize{12}{13}\selectfont\bfseries}}
\renewcommand{\subsection}{\@startsection{subsection}{2}{0mm}%
  {0.95\baselineskip}{0.45\baselineskip}%
  {\normalfont\fontsize{10.5}{11.5}\selectfont}}
\renewcommand{\subsubsection}{\@startsection{subsubsection}{3}{0mm}%
  {0.85\baselineskip}{0.35\baselineskip}%
  {\normalfont\fontsize{10}{11}\selectfont}}
\makeatother
\newcommand{\papersubhead}[1]{%
  \par\addvspace{0.75\baselineskip}%
  \noindent\underline{#1}\par
  \addvspace{0.5\baselineskip}%
}

\usepackage[hyphens]{xurl}
\usepackage{standalone}
\usepackage{import}
\usepackage{tikz}
\usepackage{amsmath}
\usepackage{xcolor}
\usepackage{tikz, float, graphicx, subcaption, mwe}
\usetikzlibrary{shapes,arrows,positioning,fit,backgrounds}
\usepackage{todonotes}
\usepackage{algorithm}
\usepackage{algpseudocode}
\usepackage{textcomp}
\usepackage{array}
\usepackage{multirow}
\usepackage{tabularx}
\usepackage{booktabs}
\usepackage{caption}
\usepackage{cuted}
\usetikzlibrary{positioning,3d,arrows.meta,shapes.geometric,backgrounds,fit,calc,matrix,decorations.pathreplacing}
\usepackage{ifthen}
\title{Towards Scaling Law Analysis For Spatiotemporal Weather Data}

\author{Alexander Kiefer$^{1,2}$, Prasanna Balaprakash$^{2}$, Xiao Wang$^{1,2}$\\[0.5em]
$^{1}$Bredesen Center, University of Tennessee, Knoxville, Tennessee\\
$^{2}$Oak Ridge National Laboratory, Oak Ridge, Tennessee\\
}

\date{} 

\begin{document}
\raggedbottom
\clearpage
\maketitle 

\thispagestyle{empty} 

\begin{abstract}
Compute-optimal scaling laws are relatively well studied for NLP and CV, where objectives are typically single-step and targets are comparatively homogeneous. Weather forecasting is harder to characterize in the same framework: autoregressive rollouts compound errors over long horizons, outputs couple many physical channels with disparate scales and predictability, and globally pooled test metrics can disagree sharply with per-channel, late-lead behavior implied by short-horizon training. We extend neural scaling analysis for autoregressive weather forecasting from single-step training loss to long rollouts and per-channel metrics. We quantify (1) how prediction error is distributed across channels and how its growth rate evolves with forecast horizon, (2) if power law scaling holds for test error, relative to rollout length when error is pooled globally, and (3) how that fit varies jointly with horizon and channel for parameter, data, and compute-based scaling axes. We find strong cross-channel and cross-horizon heterogeneity: pooled scaling can look favorable while many channels degrade at late leads. We discuss implications for weighted objectives, horizon-aware curricula, and resource allocation across outputs.
\end{abstract}

\section{Introduction}

Large neural models for scientific spatio-temporal prediction---weather and climate emulation in particular---are now trained at substantial parameter, data, and compute budgets \cite{szwarcman_prithvi-eo-20_2024,schmude_prithvi_2024,nguyen_physix_2025,subramanian_towards_2023,herde_poseidon_2024,nguyen_climax_2023,qin_timo_2025}. A practical question facing such projects is how to allocate those budgets: train a wider model, on more data, or spend more floating point operations (FLOPs) training, subject to a fixed resource envelope. For language and vision, empirical neural scaling laws---power laws linking loss to model size, data, and compute---have become the standard tool for answering that question at scale \cite{hestness_deep_2017,kaplan_scaling_2020,hoffmann_empirical_2022}

The spatio-temporal data common in scientific disciplines, such as weather, fluid dynamics, and plasma physics, differs from those benchmarks in ways that complicate a direct comparison. Quality is judged after long autoregressive rollouts, where predictions are fed back as inputs and errors compound, not from a single forward pass at the training horizon. Outputs are also high-dimensional fields with many physical channels, so scalar aggregates can hide which variables track a scaling frontier and which do not. When scaling fits are interpreted under the wrong regime, resource plans and extrapolated scaling exponents may be misleading for the error people actually care about.

Weather forecasting is a useful testbed for studying these issues: standardized reanalysis data and metrics are available \cite{rasp_weatherbench_2024,peuch_copernicus_2022}, the dynamics are challenging but well studied \cite{harris_scientific_2021}, and transformer based models are increasingly deployed in production-oriented models \cite{nguyen_climax_2023,herde_poseidon_2024,schmude_prithvi_2024,qin_timo_2025,wang_orbit-2_2025,subramanian_neural_2026}. Hardware trends, likewise, motivate understanding transformer scaling in scientific workloads, not only in natural language and computer vision.

Prior work has established neural scaling analyses for global weather emulation under a deliberately simple training objective. Subramanian et al.\ \cite{subramanian_neural_2026} trained a swin transformer on ERA5, compared continual training techniques, and fitted power-law relations and compute-optimal $(N,D)$-style allocations in the same spirit as large-language-model scaling studies. Their reported scaling summaries are based on single-step deterministic training (one forward map under mean-squared error (MSE) so that capacity--data--compute trade-offs can be measured before autoregressive effects dominate. This choice is sound for isolating pretraining scaling, but it does not answer how the same IsoFLOP family behaves when error is measured after many rollout steps on held-out years, and whether channel-wise behavior aligns with globally pooled scaling curves.

We address that gap by keeping the architecture, data splits, and IsoFLOP protocol of the preceding study and reorienting the analysis toward autoregressive test-time error. We evaluate rolled-out forecasts up to 240\,h on held-out periods, characterize how error is distributed across output channels and how its growth rate varies with lead time, and refit log--log scaling relationships between test error and three scaling covariates---parameter count, training data volume, and compute---at each horizon. Relative to single-step scaling, our proposed regime directly targets the error relevant to long-horizon forecasting and exposes heterogeneity that global summaries alone would miss.

Our paper contributes the following:
\begin{itemize}
    \item \textbf{Autoregressive scaling.} We report how the coeffiecient of determination ($R^2$) of log--log scaling laws for globally pooled test root-mean-squared error (RMSE), and the fitted scaling exponent relating error to compute, evolve with forecast horizon when models are evaluated autoregressively rather than under one-step.
    \item \textbf{Channel- and horizon-resolved analysis.} We quantify per-channel error at a fixed short lead, visualize rollout error dynamics through the time derivative of RMSE, and provide heatmaps of scaling-law $R^2$ over forecast horizon and individual output channels for parameter-, data-, and compute-based frontiers.
    \item \textbf{Evaluation protocol detail.} We connect the empirical observations---substantial cross-channel and cross-horizon heterogeneity, with strong global scaling coexisting with localized breakdowns---to practical implications such as weighted training objectives and horizon-aware objective functions.
\end{itemize}
\section{Related Work}
\label{sec:related}

\papersubhead{Neural Scaling Laws}

Neural scaling laws are empirical observations of models, often well approximated by power laws, that relate model size ($N$), training data ($D$), training compute ($C$), and predictive loss ($L$)\cite{hestness_deep_2017,kaplan_scaling_2020,hoffmann_empirical_2022}. A central use of these fits is compute-optimal allocation: for a fixed training budget $C$ (cumulative FLOPs), one seeks to minimize the achievable loss $L(N,D)$ subject to a cost model $\mathrm{FLOPs}(N,D)=C$,
\begin{equation}
    \bigl(N^*(C), D^*(C)\bigr) \in \mathop{\arg\min}_{N,\,D \,:\, \mathrm{FLOPs}(N,D)=C} L(N,D).
\end{equation}

Writing $N^*(C)$ and $D^*(C)$ for such minimizers, one typically finds

\begin{subequations}
    \label{eq:scaling}
    \begin{align}
    N^*(C) \propto C^\alpha, \label{eq:scaling-n} \\
    D^*(C) \propto C^\beta, \label{eq:scaling-d}
    \end{align}
\end{subequations}

for exponents $\alpha,\beta$ that depend on the fitted law and training setup; in the common approximation $C \propto ND$, one expects $\alpha+\beta\approx 1$ up to the details of the parameterization.

An earlier line of work on learning curves in economics and related fields studied how performance scales with data, often holding model capacity fixed \cite{mohr_learning_2024}. Hestness et al.\ \cite{hestness_deep_2017} extended this perspective with a broad study of deep networks, observing power-law relations between error or loss and both model and data scale across machine translation, language modeling, image classification, and speech recognition. A key takeaway was that the scaling exponents depend strongly on the \textbf{task}, whereas many architectural choices mainly rescale prefactors rather than the exponent.

For autoregressive transformer language models, Kaplan et al.\ \cite{kaplan_scaling_2020} fitted joint scaling of loss with parameter count $N$, token count $D$, and compute, and derived the compute-optimal bounds for their experimental regime. Their analysis implied that, at a fixed FLOP budget, performance was improved by training larger models on smaller token budgets than was the norm at the time. Hoffmann et al.\ \cite{hoffmann_empirical_2022} also explored compute-optimal training at larger scale and with more refined accounting of model size. They found a substantially more data-heavy optimum. In their setting, compute-optimal training uses many more tokens per parameter than the Kaplan study. In short, training tokens and non-embedding parameters should scale together, with on the order of twenty tokens per parameter for the one-epoch, cosine-schedule setups they emphasize. This finding, called Chinchilla scaling after their reference model, does not negate power-law scaling itself, but refines the empirically inferred balance between $N$ and $D$ at fixed $C$. Related scaling analyses have been reported for vision transformers across classification, captioning, VQA, and transfer settings \cite{alabdulmohsin_getting_2023}.

\papersubhead{Weather Forecasting}

Weather forecasting is a critical problem with real-world impact, from predicting extreme weather phenomena to quantifying actuarial risk. The objective is to predict the future state of multiple atmospheric variables that jointly describe the atmosphere.

The traditional and most trusted approach is numerical weather prediction (NWP). In NWP, the three-dimensional atmospheric state is modeled on a global or regional grid, augmented by boundary conditions and external forcings, before integrating the fluid and thermodynamic equations forward in time. The initial state is generated by data assimilation, which combines observations with a short-range prior forecast so that the model starts from physically accurate initial conditions. These systems couple large-scale dynamics, subgrid physics, and assimilation, and their computational cost grows quickly with spatial resolution, forecast length, and ensemble size \cite{harris_scientific_2021,zhao_gfdl_2018,eshagh_earth_2025}.

This expensive cycle of numerical simulation motivates neural surrogates. Instead of directly integrating the primitive equations, one learns an approximation to the mapping from recent observations to future states as a multivariate spatio-temporal regression problem.

Let $\mathbf{x}_{t}$ denote the observed state vector of atmospheric variables at time $t$. The forecasting task aims to learn a function $f_\theta$ such that

\begin{equation}
    \hat{\mathbf{x}}_{t+\tau} = f_\theta(\mathbf{x}_{t}, \mathbf{x}_{t-1}, \ldots, \mathbf{x}_{t-k+1}),
\end{equation}

where $k$ is the input history length and $\tau$ is the lead time of the forecast. The model parameters $\theta$ are learned by minimizing a loss $L$ over a dataset $\mathcal{D}$ of historical observations:

\begin{equation}
    \min_{\theta}\; \frac{1}{|\mathcal{D}|} \sum_{(\mathbf{x}_{t-k+1:t},\, \mathbf{x}_{t+\tau}) \in \mathcal{D}} L\big(f_\theta(\mathbf{x}_{t-k+1:t}),\; \mathbf{x}_{t+\tau}\big).
\end{equation}

Neural surrogate models have proven to be highly effective at trading the extremely high cost of NWP models for relatively modest penalties in accuracy \cite{bodnar_foundation_2025,holden_emulation_2015,wang_orbit-2_2025,schmude_prithvi_2024,nguyen_climax_2023,lam_learning_2023}. Additionally, these deterministic forecasting models lend themselves well to extensions as probabilistic forecasting models, where an ensemble of deterministic models with slightly varying initial conditions samples from the distribution and minimizes a scoring function, such as the continuous ranked probability score (CRPS) \cite{matheson_scoring_1976}, as in FourCastNet \cite{bonev_fourcastnet_2025}.

\begin{figure*}[t!]
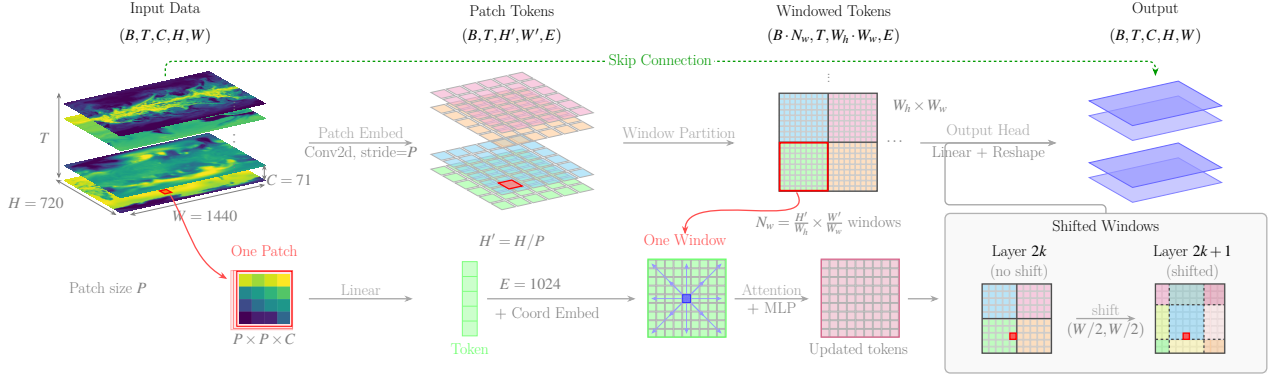

    \centering
    \includestandalone[width=\textwidth]{figures/swin_data_flow}
    \caption{Swin weather forecaster: patch embedding of the multi-channel field, stacked shifted-window Swin blocks, and patch unembedding to the predicted state.}
    \label{fig:data_flow}
\end{figure*}

\papersubhead{Scaling Laws for Weather}

Subramanian et al.\ \cite{subramanian_neural_2026} presented one of the first rigorous study of neural scaling laws for global weather emulation. They trained a family of minimal, scalable Swin Transformers on ERA5 and compared continual training with a constant learning rate and periodic cooldowns against repeated from-scratch runs under cosine decay. The continual recipe matches or improves validation loss while substantially reducing the cost of sweeping many compute budgets. Across model sizes, dataset sizes, and FLOP budgets they constructed IsoFLOP frontiers and identified compute-optimal $(N,D)$-style allocations in the same form as scaling analyses for large language models, but for a deterministic weather forecasting task.

That work is focused on single-step deterministic training, where the network maps the state at time $t$ to the state at $t+\Delta t$ under a mean-squared-error objective, so that capacity/data/compute trade-offs can be measured before compounding effects of long autoregressive rollouts dominate. We reuse the architecture, ERA5 splits, and IsoFLOP training design, but our analyses, motivated in the introduction, target autoregressive test-time error, per-channel structure, and horizon-dependent scaling diagnostics.
\section{Methodology}
\label{sec:methods}

This section isolates what is new in our paper. We reuse the architecture, ERA5 preprocessing, and IsoFLOP training family from \cite{subramanian_neural_2026}, but change the target of analysis from single-step training loss to autoregressive rollout error and its scaling behavior over lead time and output channel.

\papersubhead{Rollout Evaluation Framework}

At inference, the network is applied recursively: the prediction at $t+\Delta t$ is fed back into the input window for the next six-hour step. We evaluate on the held-out years in Table~\ref{tab:dataset_specs}. Initial conditions (ICs) are sampled every 12\,h in the test window, and each IC is rolled out to fixed lead times up to 240\,h. Metrics are logged per lead and per channel, then reduced across data-parallel workers.

\papersubhead{Rollout Scaling-law Analysis}

We reuse the same IsoFLOP frontiers and compute-optimal allocations from training, but replace single-step training loss with inference error at each rollout horizon. For each lead time $h$ (and, where applicable, channel $c$), we fit
\begin{equation}
    \log \epsilon_{h,c} = a_{h,c} + b_{h,c}\log s,
\end{equation}
where $s$ is the scaling covariate (optimal parameter count, optimal data size, or compute along the frontier) and $\epsilon$ is RMSE. The slope $b_{h,c}$ is the empirical scaling exponent at that horizon/channel, and $R^2_{h,c}$ quantifies how well a linear relation in log--log space explains the observed frontier.
This is a two-stage procedure. For each FLOP budget we first estimate an IsoFLOP optimum from an order-2 polynomial fit in log--log space, then we fit the linear power law above to those extracted optima across budgets.
For each horizon/channel fit, we report
\begin{equation}
    R^2_{h,c}
    = 1 - \frac{\sum_i \left(\log \epsilon_{h,c,i} - \widehat{\log \epsilon}_{h,c,i}\right)^2}
    {\sum_i \left(\log \epsilon_{h,c,i} - \overline{\log \epsilon}_{h,c}\right)^2},
\end{equation}
where $i$ indexes FLOP budgets (or extracted optima), $\widehat{\log \epsilon}_{h,c,i}$ is the fitted log-error, and $\overline{\log \epsilon}_{h,c}$ is the sample mean log-error for that $(h,c)$ slice.

\papersubhead{Metrics}

Unless otherwise noted, we report area-weighted RMSE per channel and lead time, where latitude weights account for grid-cell area on the latitude--longitude mesh. We also report globally pooled RMSE (the square root of channel-uniform global MSE used in training) as a single scalar baseline.

\section{Experimental Setup}
\label{sec:exp-setup}
\papersubhead{Model Architecture}

We use a swin transformer \cite{liu_swin_2021} adapted to latitude--longitude fields as in \cite{subramanian_neural_2026}: patch embedding maps the multi-channel atmospheric state to a latent grid; stacked swin blocks apply shifted-window multi-head self-attention and an MLP with pre-RMSNorm, residual connections, and QK-normalization for stability. Alternate layers shift the window partition so information mixes across windows. The mask respects east--west periodicity. A final head maps latents back to physical channels.

\papersubhead{Distributed Training}

\begin{figure}[H]
    \centering
\newcommand{\dpdist}{3cm}  
\newcommand{\sprowsep}{1.75cm}  
\newcommand{\spcolsep}{1.75cm}  
\newcommand{\tpdist}{1.25cm}   
\newcommand{\wgradbend}{25}    
\newcommand{\splabelpos}{north} 
\newcommand{\splabelxshift}{0pt}   
\newcommand{\splabelyshift}{45pt}  
\newcommand{\commsBodyDown}{1cm}

\newcommand{\commsFigBody}{\Huge}              
\newcommand{\commsFigDP}{\Huge\bfseries}             
\newcommand{\commsFigSP}{\Huge}                     
\newcommand{\commsFigBrace}{\Huge\bfseries}    
\newcommand{\commsFigFont}{\sffamily\commsFigBody}
\newcommand{\commsFigDPFont}{\sffamily\commsFigDP}
\newcommand{\commsFigSPFont}{\sffamily\commsFigSP}
\newcommand{\commsFigBraceFont}{\sffamily\commsFigBrace}

\tikzset{
    comms-dp-label/.style={yshift=1.3cm, font=\commsFigDPFont},
    comms-sp-label/.style={
        overlay,
        anchor=\splabelpos,
        xshift=\splabelxshift,
        yshift=\splabelyshift,
        font=\commsFigSPFont
    }
}

\begin{tikzpicture}[
    font=\commsFigFont,
    dp-node/.style={draw, thick, fill=blue!10, minimum width=11cm, minimum height=11cm},
    sp-node/.style={draw, fill=green!10, minimum width=4.25cm, minimum height=4.25cm, align=center},
    tp-node/.style={draw, fill=orange!20, minimum width=1.5cm, minimum height=1.5cm, align=center, inner sep=2pt},
    arrow/.style={thick,->,>=Stealth},
    brace/.style={decorate, decoration={brace, amplitude=10pt}},
    brace-vert/.style={decorate, decoration={brace, amplitude=10pt, mirror}}
]

\begin{scope}[yshift=-\commsBodyDown]
\node (dp0) [dp-node, label={[comms-dp-label]above:Data Parallel Rank 0},] {};

\matrix (sp_matrix0) [matrix of nodes, 
    nodes={sp-node}, 
    row sep=\sprowsep, column sep=\spcolsep, 
    at=(dp0.center)
]
{
    \node(sp0_0) [label={[comms-sp-label]\splabelpos:SP Rank 0}] {}; & \node(sp0_1) [label={[comms-sp-label]\splabelpos:SP Rank 1}] {}; \\
    \node(sp0_2) [label={[comms-sp-label]\splabelpos:SP Rank 2}] {}; & \node(sp0_3) [label={[comms-sp-label]\splabelpos:SP Rank 3}] {}; \\
};

\node (tp0_0) [tp-node, at=(sp0_0.center), xshift=-\tpdist] {TP 0};
\node (tp0_1) [tp-node, at=(sp0_0.center), xshift=\tpdist] {TP 1};

\node (tp1_0) [tp-node, at=(sp0_1.center), xshift=-\tpdist] {TP 0};
\node (tp1_1) [tp-node, at=(sp0_1.center), xshift=\tpdist] {TP 1};

\node (tp2_0) [tp-node, at=(sp0_2.center), xshift=-\tpdist] {TP 0};
\node (tp2_1) [tp-node, at=(sp0_2.center), xshift=\tpdist] {TP 1};

\node (tp3_0) [tp-node, at=(sp0_3.center), xshift=-\tpdist] {TP 0};
\node (tp3_1) [tp-node, at=(sp0_3.center), xshift=\tpdist] {TP 1};

\node (dp1) [dp-node, label={[comms-dp-label]above:Data Parallel Rank 1}, right=\dpdist of dp0] {};

\matrix (sp_matrix1) [matrix of nodes, 
    nodes={sp-node}, 
    row sep=\sprowsep, column sep=\spcolsep, 
    at=(dp1.center)
]
{
    \node(sp1_0) [label={[comms-sp-label]\splabelpos:SP Rank 0}] {}; & \node(sp1_1) [label={[comms-sp-label]\splabelpos:SP Rank 1}] {}; \\
    \node(sp1_2) [label={[comms-sp-label]\splabelpos:SP Rank 2}] {}; & \node(sp1_3) [label={[comms-sp-label]\splabelpos:SP Rank 3}] {}; \\
};

\node (tp4_0) [tp-node, at=(sp1_0.center), xshift=-\tpdist] {TP 0};
\node (tp4_1) [tp-node, at=(sp1_0.center), xshift=\tpdist] {TP 1};

\node (tp5_0) [tp-node, at=(sp1_1.center), xshift=-\tpdist] {TP 0};
\node (tp5_1) [tp-node, at=(sp1_1.center), xshift=\tpdist] {TP 1};

\node (tp6_0) [tp-node, at=(sp1_2.center), xshift=-\tpdist] {TP 0};
\node (tp6_1) [tp-node, at=(sp1_2.center), xshift=\tpdist] {TP 1};

\node (tp7_0) [tp-node, at=(sp1_3.center), xshift=-\tpdist] {TP 0};
\node (tp7_1) [tp-node, at=(sp1_3.center), xshift=\tpdist] {TP 1};


\draw [arrow, <->, red, line width=1.5pt] (tp0_0.east) -- (tp0_1.west);
\draw [arrow, <->, red, line width=1.5pt] (tp1_0.east) -- (tp1_1.west);
\draw [arrow, <->, red, line width=1.5pt] (tp2_0.east) -- (tp2_1.west);
\draw [arrow, <->, red, line width=1.5pt] (tp3_0.east) -- (tp3_1.west);
\draw [arrow, <->, green!50!black, dashed, bend right=\wgradbend] (tp0_0.east) to (tp1_0.west);
\draw [arrow, <->, green!50!black, dashed, bend right=\wgradbend] (tp0_1.east) to (tp1_1.west);
\draw [arrow, <->, green!50!black, dashed, bend right=\wgradbend] (tp0_0.south) to (tp2_0.north);
\draw [arrow, <->, green!50!black, dashed, bend right=\wgradbend] (tp0_1.south) to (tp2_1.north);
\draw [arrow, <->, green!50!black, dashed, bend right=\wgradbend] (tp1_0.south) to (tp3_0.north);
\draw [arrow, <->, green!50!black, dashed, bend right=\wgradbend] (tp1_1.south) to (tp3_1.north);
\draw [arrow, <->, green!50!black, dashed, bend right=\wgradbend] (tp2_0.east) to (tp3_0.west);
\draw [arrow, <->, green!50!black, dashed, bend right=\wgradbend] (tp2_1.east) to (tp3_1.west);
\draw [arrow, <->, violet!70!black, dashed] (sp0_0.east) -- (sp0_1.west);
\draw [arrow, <->, violet!70!black, dashed] (sp0_0.south) -- (sp0_2.north);
\draw [arrow, <->, violet!70!black, dashed] (sp0_1.south) -- (sp0_3.north);
\draw [arrow, <->, violet!70!black, dashed] (sp0_2.east) -- (sp0_3.west);

\draw [arrow, <->, red, line width=1.5pt] (tp4_0.east) -- (tp4_1.west);
\draw [arrow, <->, red, line width=1.5pt] (tp5_0.east) -- (tp5_1.west);
\draw [arrow, <->, red, line width=1.5pt] (tp6_0.east) -- (tp6_1.west);
\draw [arrow, <->, red, line width=1.5pt] (tp7_0.east) -- (tp7_1.west);
\draw [arrow, <->, green!50!black, dashed, bend right=\wgradbend] (tp4_0.east) to (tp5_0.west);
\draw [arrow, <->, green!50!black, dashed, bend right=\wgradbend] (tp4_1.east) to (tp5_1.west);
\draw [arrow, <->, green!50!black, dashed, bend right=\wgradbend] (tp4_0.south) to (tp6_0.north);
\draw [arrow, <->, green!50!black, dashed, bend right=\wgradbend] (tp4_1.south) to (tp6_1.north);
\draw [arrow, <->, green!50!black, dashed, bend right=\wgradbend] (tp5_0.south) to (tp7_0.north);
\draw [arrow, <->, green!50!black, dashed, bend right=\wgradbend] (tp5_1.south) to (tp7_1.north);
\draw [arrow, <->, green!50!black, dashed, bend right=\wgradbend] (tp6_0.east) to (tp7_0.west);
\draw [arrow, <->, green!50!black, dashed, bend right=\wgradbend] (tp6_1.east) to (tp7_1.west);
\draw [arrow, <->, violet!70!black, dashed] (sp1_0.east) -- (sp1_1.west);
\draw [arrow, <->, violet!70!black, dashed] (sp1_0.south) -- (sp1_2.north);
\draw [arrow, <->, violet!70!black, dashed] (sp1_1.south) -- (sp1_3.north);
\draw [arrow, <->, violet!70!black, dashed] (sp1_2.east) -- (sp1_3.west);

\draw [arrow, <->, green!50!black, dashed, bend right=\wgradbend] (dp0.east) to (dp1.west);

\draw [brace] (sp0_0.north west) -- (sp0_1.north east) node [midway, above=10pt, font=\commsFigBraceFont] {\textbf{SP2} (Width)};
\draw [brace-vert] (sp0_0.north west) -- (sp0_2.south west) node [midway, left=10pt, align=center, font=\commsFigBraceFont] {\textbf{SP1}\\(Height)};

\end{scope}

\node (legend) [
    draw,
    rounded corners,
    anchor=south,
    above=1.5cm of dp0.north, xshift=7cm,
    yshift=\commsBodyDown,
    font=\commsFigFont
] {
    \begin{tabular}{@{}c@{\hspace{0.25cm}}l@{\hspace{0.8cm}}c@{\hspace{0.25cm}}l@{\hspace{0.8cm}}c@{\hspace{0.25cm}}l@{\hspace{0.8cm}}c@{\hspace{0.25cm}}l@{\hspace{0.8cm}}c@{\hspace{0.25cm}}l@{\hspace{0.8cm}}c@{\hspace{0.25cm}}l@{}}
        \textcolor{blue!70!black}{\rule{0.25cm}{0.25cm}} & DP &
        \textcolor{green!70!black}{\rule{0.25cm}{0.25cm}} & SP &
        \textcolor{orange!70!black}{\rule{0.25cm}{0.25cm}} & TP &
        \tikz[baseline=-0.5ex]\draw[<->, red, line width=1.2pt] (0,0) -- (0.5,0); & Activations AllReduce &
        \tikz[baseline=-0.5ex]\draw[<->, green!50!black, dashed, bend right=\wgradbend] (0,0) to (0.5,0); & Wgrad AllReduce &
        \tikz[baseline=-0.5ex]\draw[<->, violet!70!black, dashed] (0,0) -- (0.5,0); & Roll
    \end{tabular}
};

\end{tikzpicture}
    \caption{Distributed communication in hybrid DP--SP training: gradient synchronization across data-parallel replicas, halo and roll traffic for shifted-window attention under spatial decomposition, and partial-sum reductions for sharded attention and MLP blocks.}
    \label{fig:comms}
\end{figure}
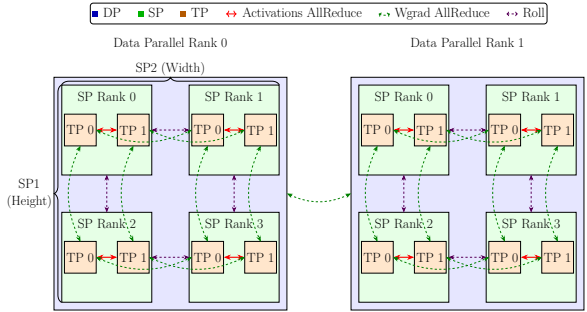

We use the hybrid data-spatial-tensor parallel distributed training framework from \cite{subramanian_neural_2026}. Global fields at $0.25^\circ$ yield very large spatial tensors. With small batch sizes, activation memory often dominates over weight storage, especially when fine-tuning or analyzing autoregressive unrolls. We therefore rely on hybrid parallel training that combines data parallel (DP) replication over batches with two-dimensional spatial domain decomposition (SP1 along latitude, SP2 along longitude). This model also supports additional decomposition dimensions (for example, depth-parallel SP3, for 3D spatial inputs); they are not required for our runs as we treat our ERA5 inputs as 2D images. Tensor parallel (TP) can shard attention heads and MLP hidden dimensions with AllReduce collectives. A typical layout uses $dp \times sp_1 \times sp_2$ GPUs (and optional $sp_3$, $tp$), with DP groups outside and SP--TP nested within. Local patch-grid extents scale inversely with $sp_1$, $sp_2$. Shifted-window attention requires halo exchanges and a distributed cyclic roll along decomposed latitude--longitude axes so each rank sees correct window boundaries---Figure~\ref{fig:comms} sketches these communication patterns.

Per-device activation footprint scales roughly as $O(B\,H_\ell W_\ell E)$ for local height--width $H_\ell,W_\ell$, batch $B$, and embed dimension $E$, so increasing $sp_1 \cdot sp_2$ reduces memory in proportion to the subdomain area until communication or surface-to-volume effects dominate. The Swin window size (in patches) further limits how far $sp_1$ and $sp_2$ can be pushed; each rank must contain atleast one window.

\papersubhead{Data}

The network takes macroscopic variables and time-invariant fields (land--sea mask, orography, cosine latitude). The model input has shape $B \times T \times C \times H \times W$, where $B$ is batch size, $T$ is context length, $C$ is the per-step channel count, and $H,W$ are spatial dimensions. The autoregressive prediction target is the next state with shape $B \times 1 \times C \times H \times W$. In our experiments, we limit the size of $T$ to $1$ across all experiments.

We use the ERA5 reanalysis \cite{hersbach_era5_2020} with the 71-channel subset and pressure levels listed in Table~\ref{tab:dataset_specs}.

\begin{table}[H]
    \centering
    \footnotesize
    \setlength{\tabcolsep}{3pt}
    \renewcommand{\arraystretch}{1.05}
    \begin{tabularx}{\columnwidth}{@{}p{0.36\columnwidth} X@{}}
    \hline
    \textbf{Property} & \textbf{Description} \\
    \hline
    \hline
    Training years & 1979--2016. \\
    \hline
    Validation year & 2017. \\
    \hline
    Test years & 2018--2022. \\
    \hline
    Spatial resolution & $0.25^{\circ} \times 0.25^{\circ}$. \\
    \hline
    Grid dimensions & $1440 \times 721$ (longitude $\times$ latitude). \\
    \hline
    Surface variables & TCWV [kg/m$^2$]; u10m, v10m [m/s]; t2m [K]; sp, msl [Pa]. \\
    \hline
    Pressure-level\newline variables & $u$, $v$ [m/s]; geopotential $z$ [m$^2$/s$^2$]; temperature $t$ [K]; specific humidity $q$ [kg/kg]. \\
    \hline
    Pressure levels (hPa) & 50, 100, 150, 200, 250, 300, 400, 500, 600, 700, 850, 925, 1000. \\
    \hline
    Total channels & 6 surface $+$ 5 fields $\times$ 13 levels $= 71$. \\
    \hline
    Temporal sampling & Hourly in ERA5; the model uses a $6$\,h forecast step (inputs and targets six hours apart). \\
    \hline
    \end{tabularx}
    \caption{ERA5 setup and data splits.}
    \label{tab:dataset_specs}
\end{table}

\papersubhead{Training Protocol}

Our IsoFLOP construction matches \cite{subramanian_neural_2026}. Models are trained to predict the next six-hour state under uniform MSE over channels, with normalization statistics computed from the training split. Different IsoFLOP points are trained through constant learning rate (LR) schedule with periodic cooldowns to multiple compute budgets.

\section{Results}
\label{sec:results}

\papersubhead{Six-hour per-channel error structure}

\begin{figure}[H]
    \centering
    \includegraphics[width=\linewidth]{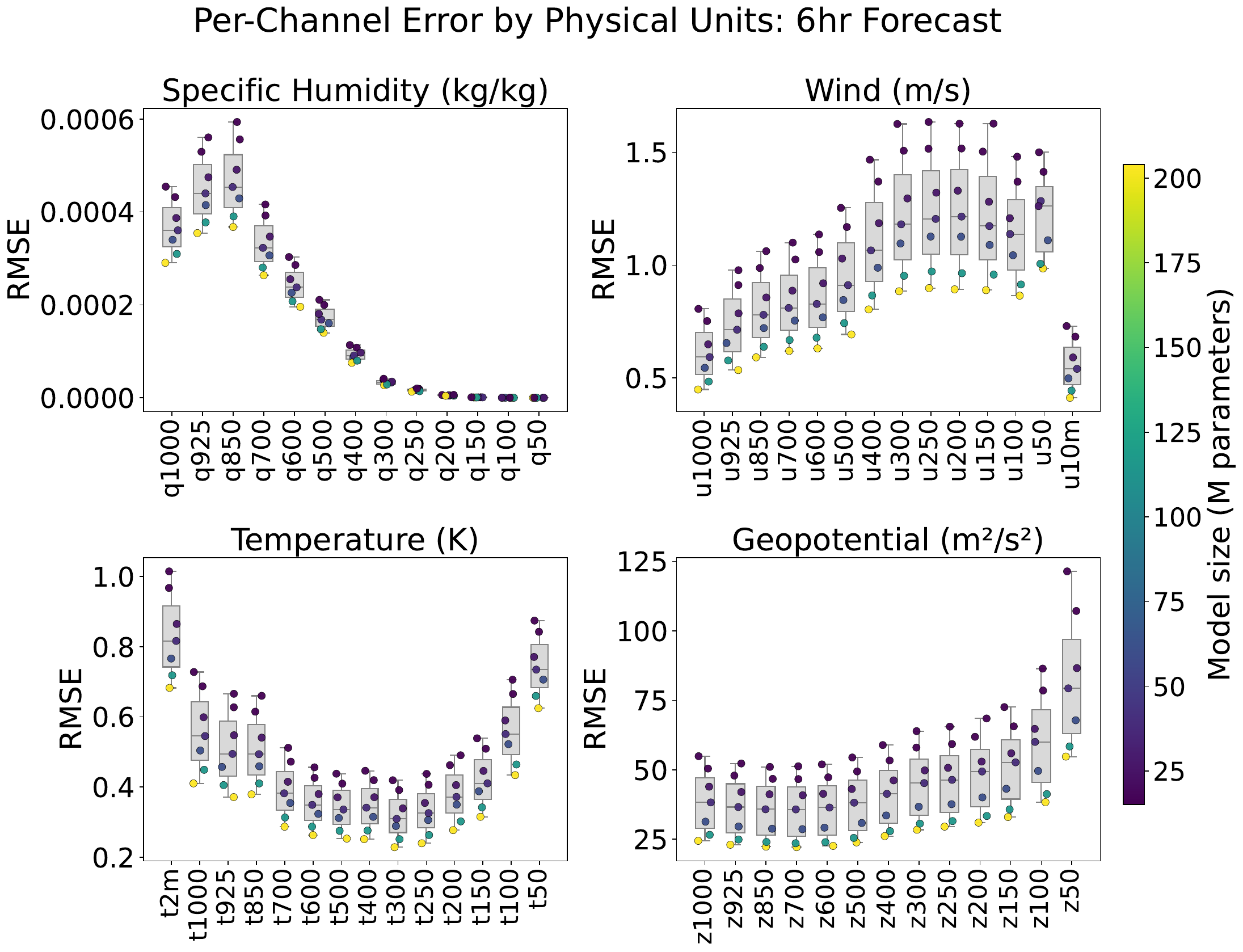}
    \caption{Per-channel area-weighted RMSE at six hours. Panels group channels by unit; each channel shows a box plot and one RMSE value per IC}
    \label{fig:rmse-dist-6h}
\end{figure}

\begin{figure*}[t]
    \centering
    \includegraphics[width=\textwidth]{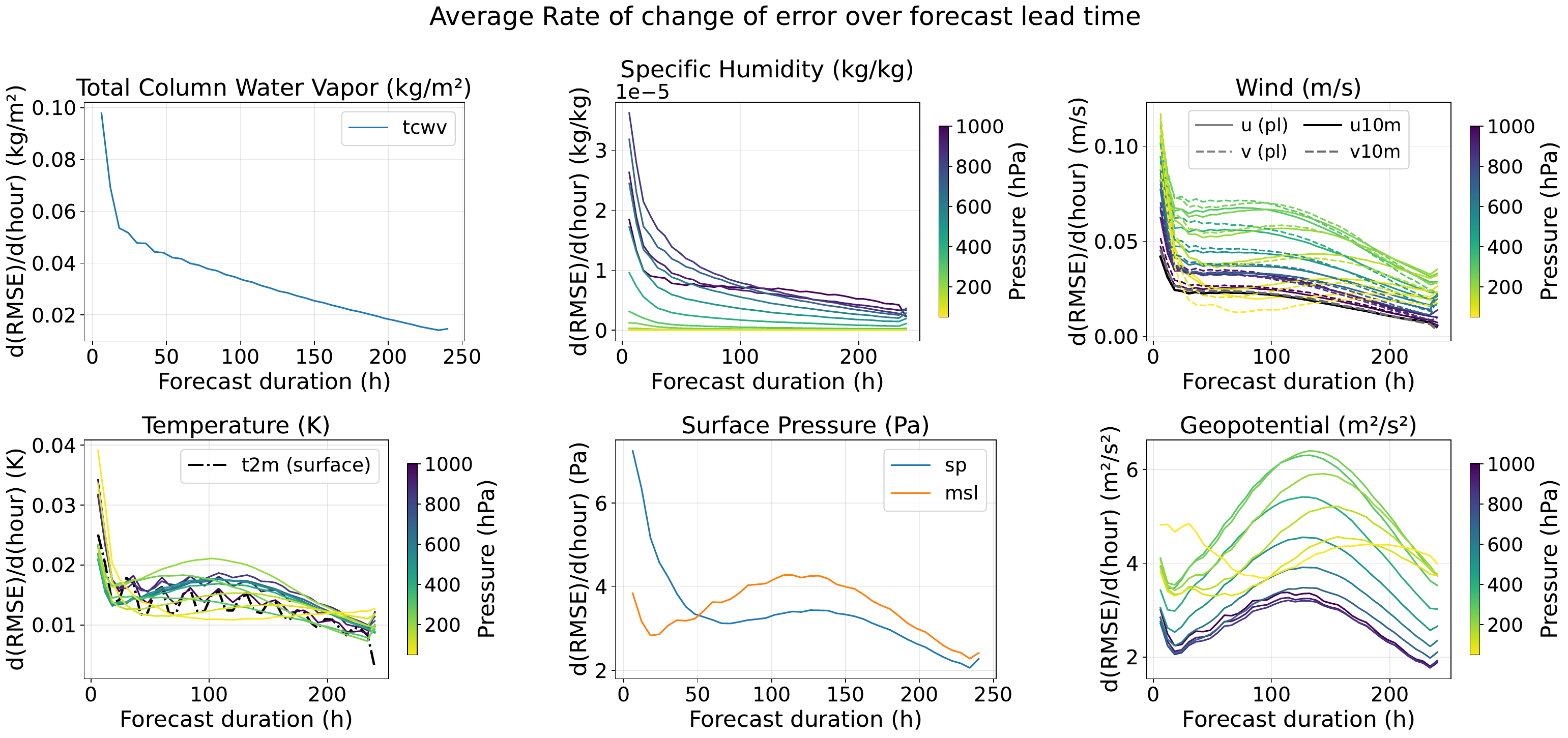}
    \caption{Average time derivative of area-weighted RMSE, $\mathrm{d}(\mathrm{RMSE})/\mathrm{d}t$, versus forecast lead time (hours), through 240\,h. Curves are averaged over ICs; panels group channels by unit (color encodes pressure level where applicable).}
    \label{fig:error-growth}
\end{figure*}

Figure~\ref{fig:rmse-dist-6h} shows how prediction error is distributed across targets at the first rollout step. Each component of the 71-dimensional output is one scalar channel; we plot area-weighted RMSE per channel, with one point per IC and a box summary on top. Channels are split into six panels by measurement unit so that values within a panel live on comparable numeric scales.

Several patterns are useful for downstream analysis. First, error is not evenly spread across channels. Even with a single global MSE loss during training, the implied per-channel RMSE ranges differ widely (both medians and spreads). Second, for variables repeated at multiple vertical indices (pressure levels), error often changes smoothly along that index: it is not i.i.d.\ across levels. For example, humidity channels tend to show larger RMSE at low levels and smaller RMSE at high levels, with a visible change of slope partway up the stack. Wind components show RMSE that grows toward the middle of the vertical index, then tapers slightly at the top. Temperature error is high for the dedicated surface channel, smaller through a band of mid-level channels, and rises again for the highest (sparsest) level we retain. Geopotential is relatively stable across most indexed levels and increases only toward the top of the stack. Among a few non-level channels in the same unit family, one pressure-related surface field is much harder than another (wider boxes and more scatter across ICs).

In short, the model induces a structured error profile over its output vector: targets differ in both typical error and IC-to-IC variability. That heterogeneity matters for any scaling-law fit that is computed per channel or pooled across channels, which we examine next.

\papersubhead{Error Growth Over Rollout}

Figure~\ref{fig:error-growth} summarizes the rate of error accumulation, not the error level: the vertical axis is the derivative of RMSE with respect to time. When this derivative is positive, RMSE is still rising; when the curve falls, the rise is slowing (error growth is decelerating), even if RMSE itself has not peaked within the window.

RMSE at a chosen lead is only a snapshot: the same value can arise from different temporal paths, and comparing two leads does not say how aggressively error is still being added in between. The derivative separates regime---whether each extra hour is still buying a large increment of error (high derivative), or whether the system is entering a saturation-like phase where RMSE may rise only slowly (small derivative). It also surfaces short windows where growth briefly speeds up again, which a few sparse RMSE evaluations can miss.

Across targets, the derivative curves are not simple translations of one another. Many channels show a rapid drop in the first day of integration: the largest incremental RMSE gains occur early, after which additional hours add error more slowly. That pattern is clearest for the single column-integrated target in the first panel and for several families of indexed channels, but the initial magnitude and decay rate differ by target.

For multi-level fields, the ordering of curves often mirrors the static RMSE stratification from Fig.~\ref{fig:rmse-dist-6h}: channels that were ``hard'' at six hours also tend to carry larger short-horizon derivatives, while some upper-level traces sit close to zero for long stretches (error barely climbs per hour). Wind and temperature panels illustrate that mid-stack indices can dominate both the level and the shape of the derivative, including modest mid-rollout structure (plateaus or secondary bumps) rather than monotone exponential decay.

The two surface pressure variables are a useful sanity check that the dynamics are target-specific: their derivative traces diverge in shape---one settles toward an approximate plateau after roughly four days while the other shows a broad mid-rollout maximum before easing. Geopotential is the starkest departure from an ``early shock then decay'': derivatives rise into a mid-horizon hump and only then decline, so error growth accelerates before it slows. That non-monotone velocity means a single functional form for ``rollout error vs.\ time'' will not fit all channels equally well.

Overall, Fig.~\ref{fig:error-growth} argues that autoregressive error should be analyzed as a multivariate time series over lead time: pooling channels or horizons without accounting for these profiles can hide which targets drive long-horizon drift. The next subsection holds the forecast horizon fixed at many values and asks how well a single power-law description of compute versus error still holds when error is measured after longer rollouts.

\papersubhead{Global scaling-law fit vs.\ forecast horizon}

\begin{figure}[H]
    \centering
    \begin{subfigure}{0.49\linewidth}
        \centering
        \includegraphics[width=\linewidth]{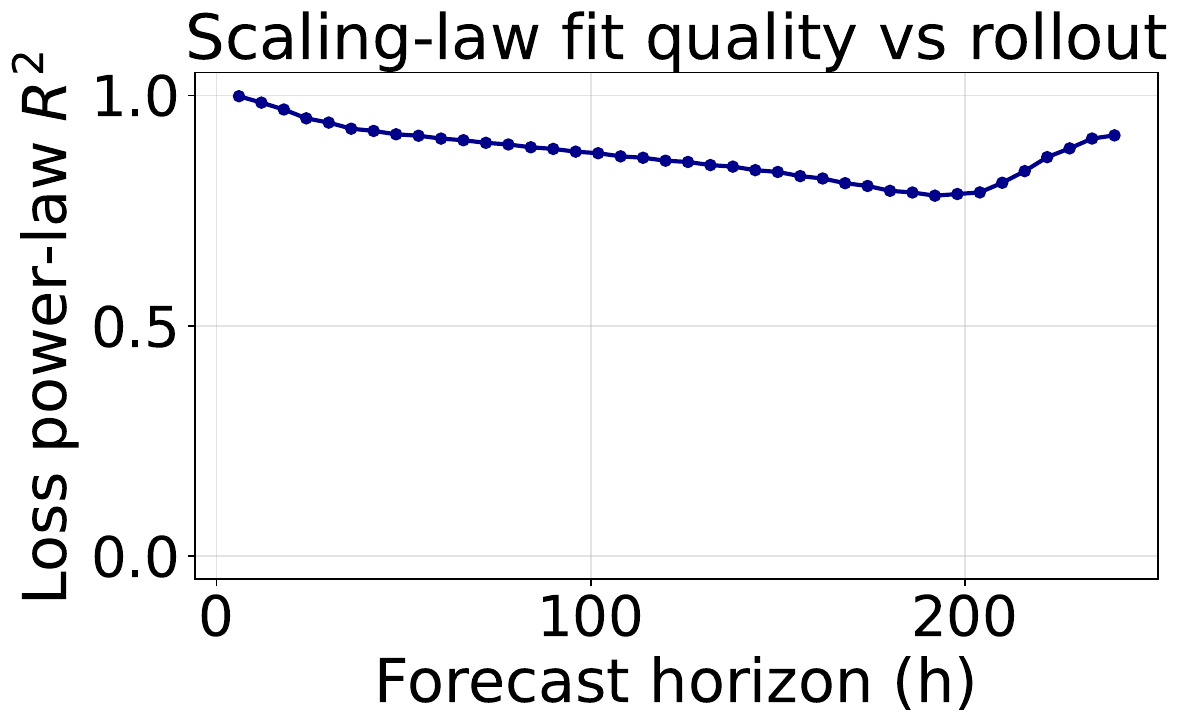}
        \phantomsubcaption
        \label{fig:global-rollout-scaling-r2}
    \end{subfigure}
    \hfill
    \begin{subfigure}{0.49\linewidth}
        \centering
        \includegraphics[width=\linewidth]{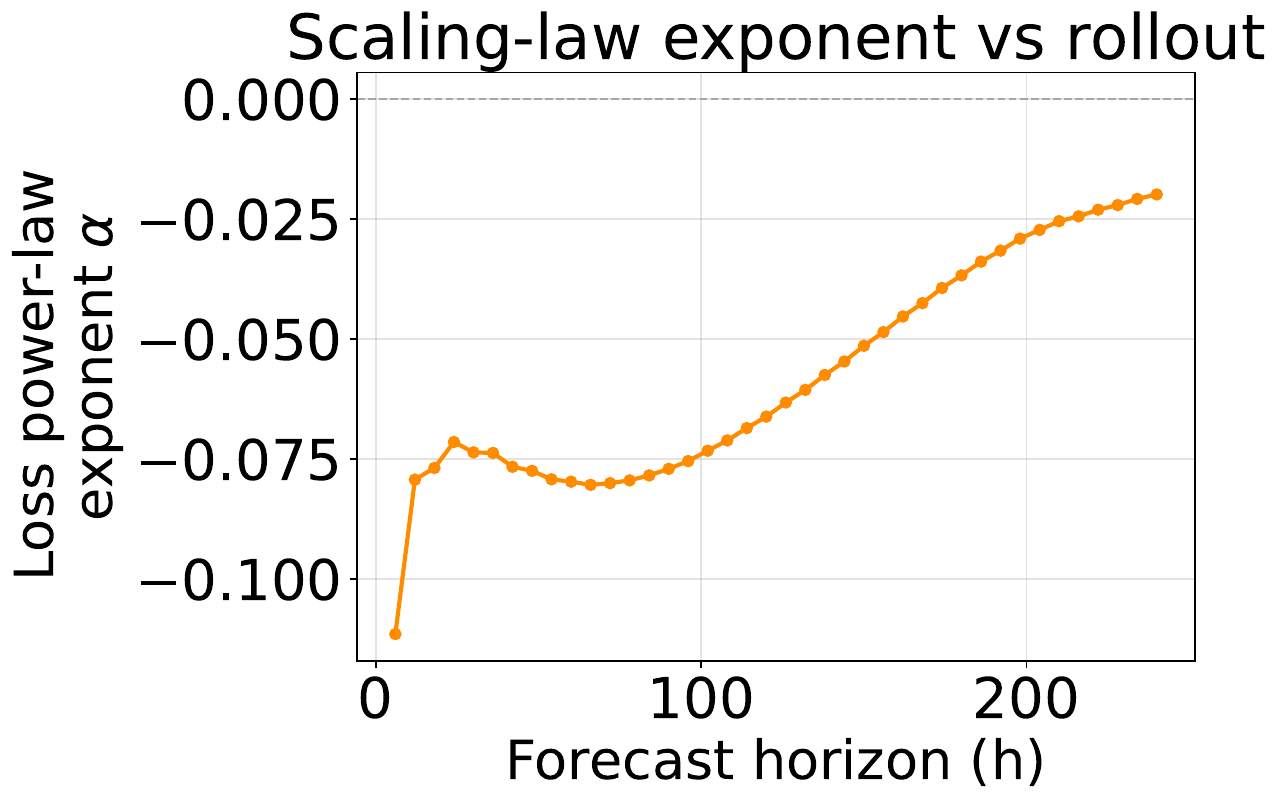}
        \phantomsubcaption
        \label{fig:global-rollout-scaling-exponent}
    \end{subfigure}
    \caption{Global scaling-law diagnostics for log--log fits of compute vs. global test RMSE, plotted as a function of forecast horizon (hours). (\subref{fig:global-rollout-scaling-r2}) shows the coefficient of determination $R^2$ as in the original analysis; (\subref{fig:global-rollout-scaling-exponent}) shows the fitted scaling exponent for each horizon. Each point re-fits the frontier using the same IsoFLOP design as in training.}
    \label{fig:global-rollout-scaling}
\end{figure}

Figure~\ref{fig:global-rollout-scaling} does not plot RMSE itself. Panel~(\subref{fig:global-rollout-scaling-r2}) shows how well a straight line in log--log space explains the relationship between compute and globally test RMSE when the latter is evaluated at each forecast horizon. Panel~(\subref{fig:global-rollout-scaling-exponent}) shows the slope of that same fit---the empirical scaling exponent linking log-compute to log-error at each horizon. Together, they explain ``is the frontier still a line?'' and ``how steep is the frontier?''

Interpreting $R^2$ as linearity after log transforms, panel~(\subref{fig:global-rollout-scaling-r2}) shows that loss-based scaling is an excellent local approximation at short leads ($R^2$ near one at the first step) but not uniformly tight across the rollout. $R^2$ drifts down through a mid-window, reaches a trough near 190--200\,h, and then rises again toward 240\,h. We do not over-interpret the trough. It is consistent with a shift in which error components dominate overall RMSE after many autoregressive steps, and with the non-monotone error-velocity patterns in Fig.~\ref{fig:error-growth}.

Panel~(\subref{fig:global-rollout-scaling-exponent}) shows complementary information about sensitivity. The fitted exponent is most negative at the shortest horizons and drifts toward zero as the evaluation lead grows, with modest non-monotone structure in the first day or two before a steadier approach to the origin at long leads. In the usual sign convention for ``more compute $\rightarrow$ lower RMSE,'' a slope closer to zero means that each additional factor of compute buys a smaller proportional reduction in error at the IsoFLOP frontier: scaling remains meaningful, but its apparent strength weakens once rollouts are long. Together with panel~(\subref{fig:global-rollout-scaling-r2}), the mid-horizon band is where linearity is worst and the slope is still far from its long-lead value. The late rise in $R^2$ does not reverse that trend toward a flatter exponent---by 240\,h the exponent is still much closer to zero than at six hours---so late-rollout error is both relatively easier to summarize with a line and relatively less responsive to compute.

\begin{figure}[H]
    \centering
    \includegraphics[width=\linewidth]{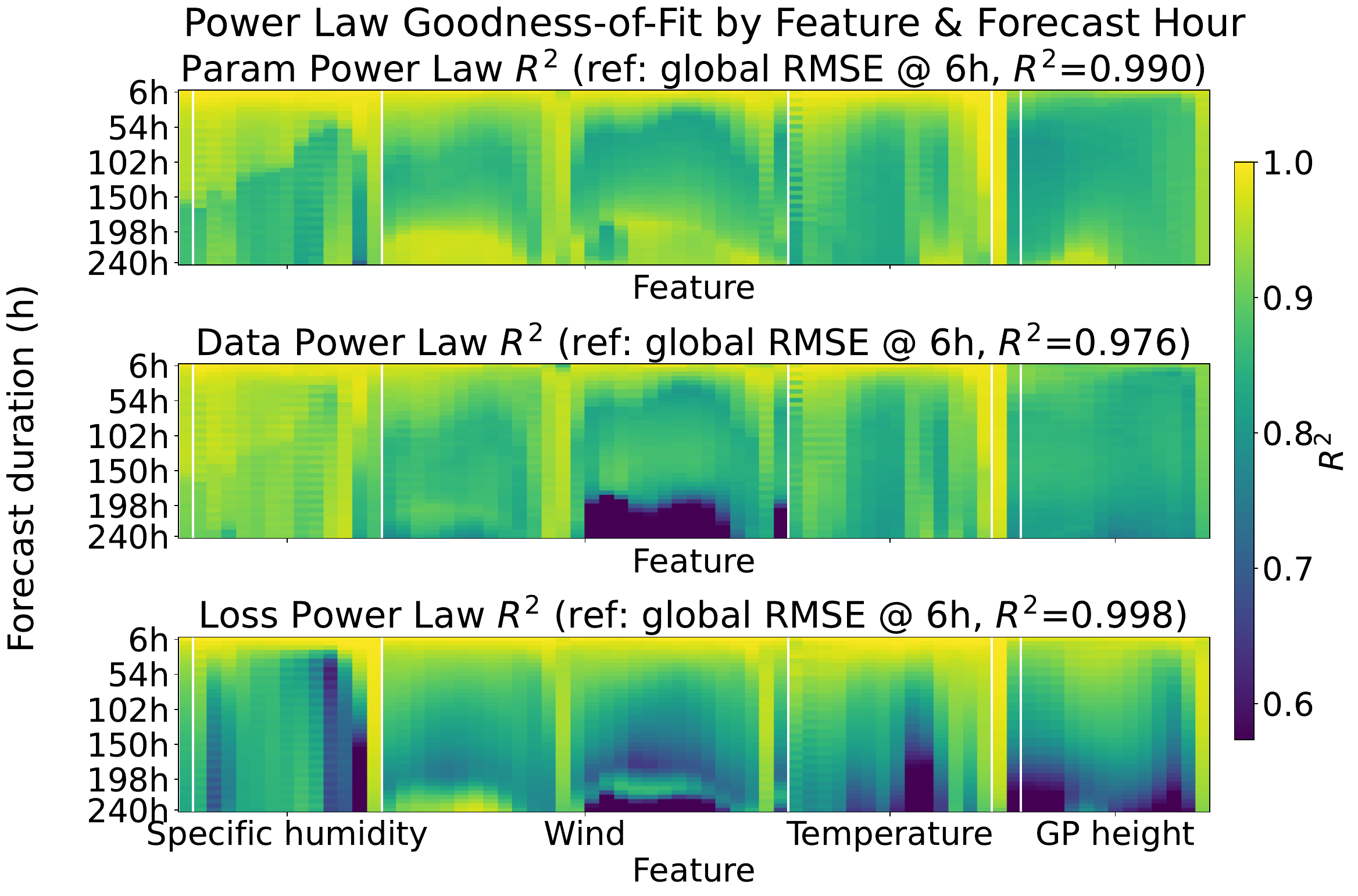}
    \caption{Heatmaps of $R^2$ for per-channel log--log scaling-law fits. Top: error vs.\ model size; middle: error vs.\ training data; bottom: error vs.\ compute. Brighter colors indicate a tighter linear fit in log--log space. Annotations give global test RMSE $R^2$ at 6\,h for each scaling axis.}
    \label{fig:per-channel-r2-heatmap}
\end{figure}

The global panels are coarse summaries: they average over all channels through pooled RMSE. Figure~\ref{fig:per-channel-r2-heatmap} resolves log--log goodness-of-fit along both forecast horizon and individual output channel, and repeats the analysis for three scaling covariates from the IsoFLOP design (parameter count, training data, and compute).

\papersubhead{Per-channel $R^2$ across horizons and scaling axes}

Figure~\ref{fig:per-channel-r2-heatmap} turns each cell of the tensor ``(channel, lead, scaling variable)'' into a single diagnostic. Reading down a column shows how well a fixed power-law survives as the evaluation rollout lengthens; reading across columns at a fixed lead shows which targets are easy or hard to describe with the same exponent family as the rest of the frontier.

The three rows answer different practical questions. \textbf{Parameter scaling} (top) is the most forgiving: $R^2$ stays high for most channels and leads, with only modest cooling (lower $R^2$) for some humidity and meridional-wind columns after roughly one week of integration. \textbf{Data scaling} (middle) is broadly similar but develops a sharper failure mode---a contiguous block of poor fits for low-level meridional wind past ${\sim}180$\,h---showing that ``more data'' and ``larger model'' are not interchangeable as predictors of long-horizon error once channels are separated. \textbf{Compute / loss scaling} (bottom) tracks the training-aligned budget most directly: it yields the strongest short-horizon linearity (consistent with the strong linearity of the global reference at six hours) but also the most heterogeneous long-horizon map, with isolated pockets of low $R^2$ for upper-level humidity, near-surface meridional wind, selected mid-to-upper temperature levels, and several geopotential columns at the longest leads.

A scalar scaling law fit to pooled loss (Fig.~\ref{fig:global-rollout-scaling}) can look excellent even while many individual channels violate the same template at late leads. The parametric row is a better ``single exponent for everyone'' summary if robustness across channels matters; the compute row is the right diagnostic if the scientific claim is tied to FLOPs or training budget. None of the three rows is uniformly dominant, which is expected once autoregressive evaluation mixes targets with different error velocities (Fig.~\ref{fig:error-growth}) and different static difficulty (Fig.~\ref{fig:rmse-dist-6h}).

\section{Discussion and Conclusions}
\label{sec:conclusions}

We explore neural scaling for a global autoregressive weather transformer, specifically in the rollout regime. The primary message of our work is that scaling behavior depends strongly on what is being measured and when it is measured. A globally pooled metric can still exhibit strong log--log structure over compute at many horizons, yet channel-resolved diagnostics reveal substantial heterogeneity in both error magnitude and scaling-law fit quality.

From a modeling perspective, this matters because long-horizon forecasting quality is not governed by a single homogeneous error process. Per-channel RMSE at short lead is already structured across variables and pressure levels, and the derivative analysis shows that error accumulation rates also differ across targets over rollout time. These differences help explain why a single exponent summary can be informative at the aggregate level but incomplete for variable-level conclusions, especially at later horizons where some channels deviate from a common frontier while others remain well behaved.

The horizon-dependent global exponent provides a complementary interpretation to $R^2$: as rollout length increases, the fitted loss--compute slope moves closer to zero, suggesting weaker proportional returns from additional compute on pooled error at long leads, even when a log--log line remains a reasonable summary. In this sense, fit quality and scaling strength are related but distinct diagnostics, and both are needed for robust conclusions about compute allocation in autoregressive scientific forecasting.

These findings suggest practical directions for training and evaluation. First, weighted or channel-aware objectives may better align optimization with long-horizon error sensitivities than uniform MSE alone. Second, horizon-aware curricula or multi-objective training could reduce the mismatch between short-horizon fit quality and late-rollout robustness. Third, future scaling studies should report both global and channel-resolved diagnostics to avoid over-generalizing from pooled metrics.

This study has limitations. Our conclusions are conditioned on the current model family, data splits, and IsoFLOP design. Extending the same analysis to alternative architectures, larger compute regimes, and probabilistic objectives is a natural next step. Overall, the results support a nuanced view: scaling laws remain useful in autoregressive weather emulation, but their reliability and practical meaning are strongly horizon and channel-dependent.
\section{Acknowledgments} 
This manuscript was co-authored by Oak Ridge National Laboratory (ORNL), operated by UT-Battelle, LLC under Contract No. DE-AC05-00OR22725 with the U.S. Department of Energy. Any subjective views or opinions expressed in this paper do not necessarily represent those of the U.S. Department of Energy or the United States Government.

This research used the Perlmutter supercomputing resources of the National Energy Research Scientific Computing Center (NERSC), a U.S. Department of Energy Office of Science User Facility located at Lawrence Berkeley National Laboratory, operated under Contract No. DE-AC02-05CH11231.
An award of computer time was provided by the ASCR Leadership Computing Challenge (ALCC) program at NERSC under ERCAP0038267.
Our conclusions do not necessarily reflect the position or the policy of our sponsors, and no official endorsement should be~inferred.



\bibliographystyle{IEEEtran}
\bibliography{bibliography}

\newpage


\begin{biography}
Alex Kiefer is a PhD student at the University of Tennessee - Knoxville and graduate research assistant at Oak Ridge National Laboratory. He received his BS and MS in computer science in 2022 from Indiana University. His work focuses on the efficient scaling of AI models on HPCs for spatio-temporal data. He is a member of ACM and SIGHPC.

Dr. Prasanna Balaprakash is the Director of AI at Oak Ridge National Laboratory. He leads research efforts at Oak Ridge in launching transformational AI model consortium (ModCon), which is a basis for DOE Genesis Mission to accelerate science through AI. He received his PhD in 2009 from the Université libre de Bruxelles.

Dr. Xiao Wang is a research staff scientist in the Computational Science and Engineering Division at Oak Ridge National Laboratory (ORNL). He holds dual Bachelor's degrees in Mathematics and Computer Science, graduating with honors from Saint John's University, MN, in 2012. He completed his M.S. and Ph.D. in Electrical and Computer Engineering at Purdue University in 2016 and 2017, under the mentorship of Dr. Charles Bouman and Dr. Samuel Midkiff. Following his Ph.D., Dr. Wang pursued postdoctoral research at Harvard Medical School and Boston Children's Hospital, specializing in medical imaging, before joining ORNL in 2021.
Dr. Wang's research focuses on the intersection of artificial intelligence (AI), high-performance computing (HPC), and computational imaging.
\end{biography}




\end{document}